\documentclass[conference]{IEEEtran}
\IEEEoverridecommandlockouts
\usepackage{cite}
\usepackage{amsmath,amssymb,amsfonts}
\usepackage{algorithmic}
\usepackage{graphicx}
\usepackage{textcomp}
\usepackage{xcolor}
\usepackage{multirow}

\def\BibTeX{{\rm B\kern-.05em{\sc i\kern-.025em b}\kern-.08em
    T\kern-.1667em\lower.7ex\hbox{E}\kern-.125emX}}
\begin{document}

\title{WeedVision: Multi-Stage Growth and Classification of Weeds using DETR and RetinaNet for Precision Agriculture\\
}

\author{
\IEEEauthorblockN{Taminul Islam}
\IEEEauthorblockA{\textit{School of Computing} \\
\textit{Southern Illinois University}\\
Carbondale, USA\\
taminul.islam@siu.edu}
\and
\IEEEauthorblockN{Toqi Tahamid Sarker}
\IEEEauthorblockA{\textit{School of Computing} \\
\textit{Southern Illinois University}\\
Carbondale, USA\\
toqitahamid.sarker@siu.edu}
\and
\IEEEauthorblockN{Khaled R Ahmed}
\IEEEauthorblockA{\textit{School of Computing} \\
\textit{Southern Illinois University}\\
Carbondale, USA\\
khaled.ahmed@siu.edu}
\and
\IEEEauthorblockN{Cristiana Bernardi Rankrape}
\IEEEauthorblockA{\textit{Department of Plant Soils and Agricultural Systems} \\
\textit{Southern Illinois University}\\
Carbondale, USA\\
cris.rankrape@siu.edu}
\and
\IEEEauthorblockN{Karla Gage}
\IEEEauthorblockA{\textit{School of Agricultural Sciences / School of Biological Sciences} \\
\textit{Southern Illinois University}\\
Carbondale, USA\\
kgage@siu.edu}
}

\maketitle

\begin{abstract}
Weed management remains a critical challenge in agriculture, where weeds compete with crops for essential resources, leading to significant yield losses. Accurate detection of weeds at various growth stages is crucial for effective management yet challenging for farmers, as it requires identifying different species at multiple growth phases. This research addresses these challenges by utilizing advanced object detection models—specifically, the Detection Transformer (DETR) with a ResNet-50 backbone and RetinaNet with a ResNeXt-101 backbone—to identify and classify 16 weed species of economic concern across 174 classes, spanning their 11-week growth stages from seedling to maturity. A robust dataset comprising 203,567 images was developed, meticulously labeled by species and growth stage. The models were rigorously trained and evaluated, with RetinaNet demonstrating superior performance, achieving a mean Average Precision (mAP) of 0.907 on the training set and 0.904 on the test set, compared to DETR's mAP of 0.854 and 0.840, respectively. RetinaNet also outperformed DETR in recall and inference speed of 7.28 FPS, making it more suitable for real-time applications. Both models showed improved accuracy as plants matured. This research provides crucial insights for developing precise, sustainable, and automated weed management strategies, paving the way for real-time species-specific detection systems and advancing AI-assisted agriculture through continued innovation in model development and early detection accuracy.
\end{abstract}


\section{Introduction}
In the vast agricultural landscape of the USA, weed management remains a critical challenge for farmers and agronomists. The diverse climates and fertile soils ideal for crop production also create favorable conditions for a wide variety of weed species \cite{monteiro2022sustainable}. These unwanted plants compete with crops for essential resources such as water, nutrients, and sunlight, potentially leading to significant yield losses and economic setbacks for farmers. Traditional weed control methods often rely on broad-spectrum herbicides or mechanically or labor-intensive removal \cite{ren2024exploring}. However, these approaches can be environmentally harmful, economically inefficient, and increasingly ineffective due to the development of herbicide-resistant weed populations \cite{Gao2024-km}. As such, there is a growing need for more precise, sustainable, and automated weed management strategies.

Recent advancements in computer vision and deep learning have shown promise in addressing this agricultural challenge. Object detection and classification techniques applied to weed identification offer the potential for highly accurate, real-time weed management solutions \cite{Almalky2023-ot}. However, several research gaps persist in this domain, such as (a) limited datasets: most existing studies rely on small datasets or images captured at specific growth stages, failing to capture the dynamic nature of weed development, and (b) lack of diversity: many datasets focus on a limited number of weed species, not reflecting the full range of weeds farmers encounter in real-world scenarios.

The scope of this work addresses these gaps by focusing on 16 weed species of greatest economic concern found commonly across multiple geographies in USA agriculture, tracking their growth from the seedling stage through 11 weeks of development. We created a robust, diverse dataset and implemented advanced object detection models to improve the accuracy and efficiency of weed identification and classification.

\begin{figure}[t]
    \centering
\includegraphics[width=0.5\textwidth,height=5cm]{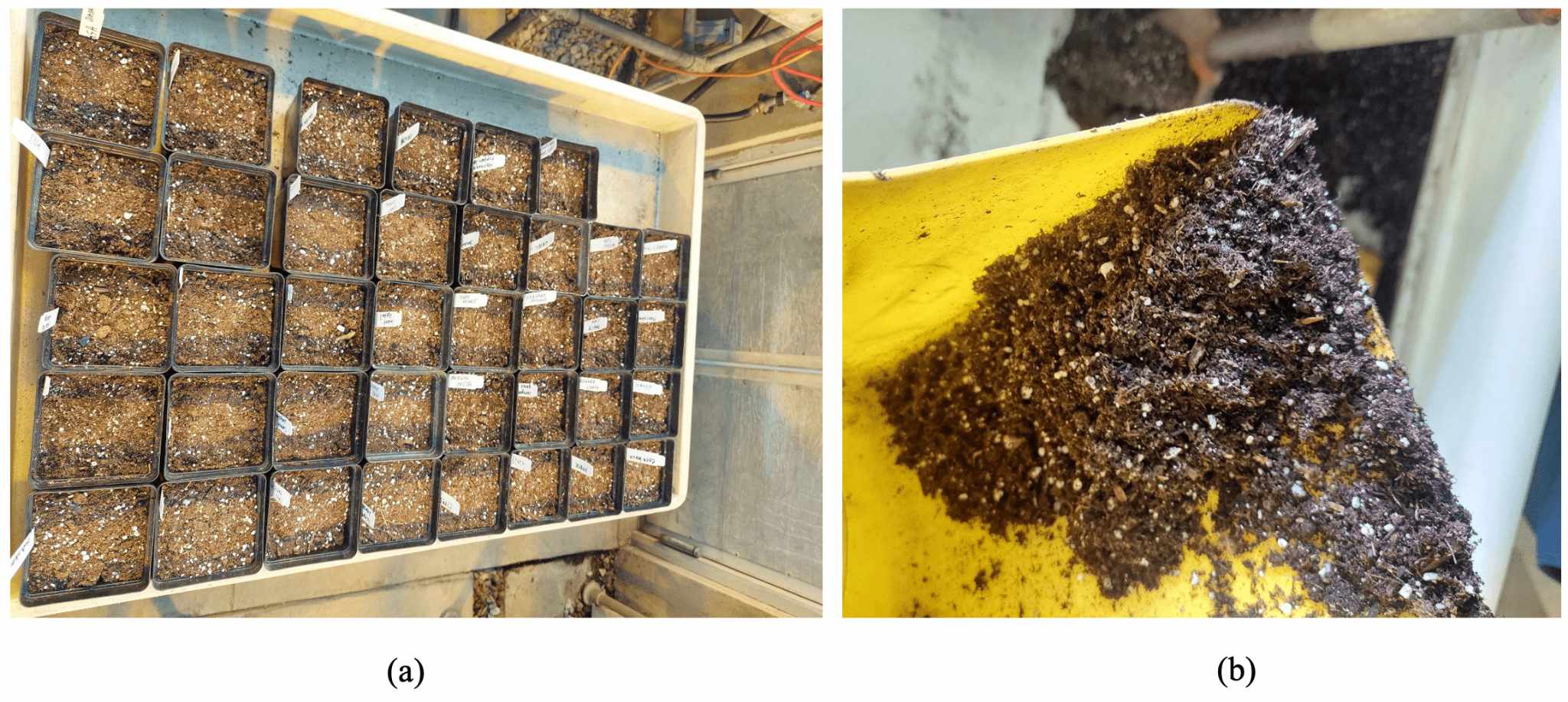}
    \vspace{-0.4cm}
    \caption{Soil preparation and labeling for planting weed seeds in pots inside the greenhouse. (a) shows the prepared pots with soil and pot stakes, (b) displays the close-up of the soil mix used for planting.}
    \label{fig:fig1}
    \vspace{-0.3cm}
\end{figure}

Our research makes several key contributions to the field:
\begin{itemize}
    \item Creation of a unique dataset comprising 203,567 images, capturing the full growth cycle of 16 of the most common and troublesome weed species in USA agriculture.
    \item Meticulous labeling of the dataset, categorized by species and growth stage (week-wise), providing a comprehensive resource for weed identification research.
    \item Implementing the Detection Transformer (DETR) \cite{Carion2020-qn} and RetinaNet \cite{Li2020-lt}, adapting these state-of-the-art object detection architectures for weed identification.
    \item Comprehensive comparison of model results, culminating in evidence-based recommendations for farmers on the most effective model for weed detection in real-world scenarios.
\end{itemize}

This research utilizes DETR and RetinaNet due to their state-of-the-art performance in object detection tasks. DETR introduces a state-of-the-art transformer-based approach, offering end-to-end object detection with potential benefits in handling complex scenes and object relationships. RetinaNet, known for its efficiency and accuracy, employs a focal loss function to address class imbalance issues common in detection tasks. By implementing and comparing these two advanced models, the study aims to evaluate their effectiveness in the specific context of weed detection and classification. This research not only contributes to the growing body of work on AI-assisted agriculture but also provides practical insights for farmers and beyond. By developing more accurate and efficient weed detection systems, we pave the way for precision agriculture techniques that can significantly reduce herbicide use, lower production costs, and minimize environmental impact.

In the following sections, this paper presents related work, followed by a comprehensive outline of the data collection and pre-processing techniques employed. The methodology section describes the steps taken in this research. Subsequently, the models section introduces the implementation and evaluation of DETR and RetinaNet for detecting and classifying 16 weed species at various growth stages. The results section showcases the performance metrics of these models. In conclusion, it summarizes these research findings for the 16 growth stage detection and classification with actionable recommendations for farmers based on the study’s
outcomes.

\section{Related Work}

Recent advancements in deep learning and computer vision have revolutionized weed detection and classification in precision agriculture. Researchers have developed various approaches to address the challenges associated with accurate and efficient weed identification in diverse crop environments.
Object detection models have shown promising results in weed identification. Hasan et al. (2024) \cite{Hasan2024-su} created a dataset of 5,997 images featuring corn and four weed species, demonstrating that YOLOv7 achieved the highest mean average precision (mAP) of 88.50\%, further improved to 89.93\% with data augmentation. Wang et al. (2024) \cite{Wang2024-tt} proposed the CSCW-YOLOv7 model for weed detection in wheat fields, achieving superior precision (97.7\%), recall (98\%), and mAP (94.4\%).
Transfer learning has proven effective for weed species detection. Shackleton et al. (2024) \cite{Shackleton2024-uc} evaluated seven pre-trained CNN models for rangeland weed detection, with EfficientNetV2B1 achieving the highest accuracy of 94.2\%. Ahmad et al. (2021) \cite{Ahmad2021-gs} employed various models for image classification and object detection in corn and soybean systems, with VGG16 achieving 98.9\% accuracy and YOLOv3 reaching 54.3\% mAP.
Traditional machine learning algorithms have also been applied to weed detection. Islam et al. (2021) \cite{Islam2021-iy} compared Random Forest, Support Vector Machine, and k-Nearest Neighbors for weed detection in chilli pepper fields, with RF and SVM achieving 96\% and 94\% accuracy, respectively.
Semantic segmentation approaches have shown promise. Khan et al. (2020) \cite{Khan2020-xb} introduced CED-Net, outperforming traditional models like U-Net and SegNet. Arun et al. (2020) \cite{Arun2020-eu} developed a Reduced U-Net architecture, achieving 95.34\% segmentation accuracy on the CWFID dataset.
Autonomous weeding applications have benefited from deep learning. Adhikari et al. (2019) \cite{Adhikari2019-os} proposed ESNet for autonomous weeding in rice fields, utilizing semantic graphics for data annotation. Teimouri et al. (2018) \cite{Teimouri2018-mx} developed a method to classify weeds into nine growth stages, achieving a maximum accuracy of 78\% for Polygonum spp.
Ensemble learning frameworks have been introduced to improve detection under varied field conditions. Asad et al. (2023) \cite{Asad2023-zv} proposed an approach using diverse models in a teacher-student configuration, significantly outperforming single semantic segmentation models. Moldvai et al. (2024) \cite{Moldvai2024-hb} explored weed detection using multiple features and classifiers, achieving a 94.56\% recall rate with limited data.
\\
Despite these advancements, several limitations persist in existing research. These include the need for larger and more diverse datasets \cite{Gallo2023-zs} \cite{Asad2020-wj}, class imbalance issues \cite{Hasan2024-su}, and computational complexity \cite{Asad2023-zv}. Most studies focus on a limited number of weed species \cite{Moldvai2024-hb} \cite{Ahmad2021-gs} and growth stages, which may not fully represent real-world agricultural settings.
Our research addresses these limitations by creating a comprehensive dataset of 203,567 images featuring 16 common and troublesome weeds in USA agriculture, capturing their full 11-week growth cycle. We implement and adapt state-of-the-art object detection models, DETR and RetinaNet, for weed growth identification. Through a comprehensive comparison of model results, we provide practical insights for weed management in precision agriculture.
This work distinguishes itself by focusing on a large-scale, diverse dataset, considering multiple weed species and growth stages, and offering practical recommendations for farmers on the most effective model for weed detection in real-world scenarios. By addressing the limitations of previous studies, our research contributes significantly to the field of weed identification and management in precision agriculture.
\section{Data Description and Preproceccing}

\begin{figure}[t]
    \centering
    \includegraphics[width=0.5\textwidth]{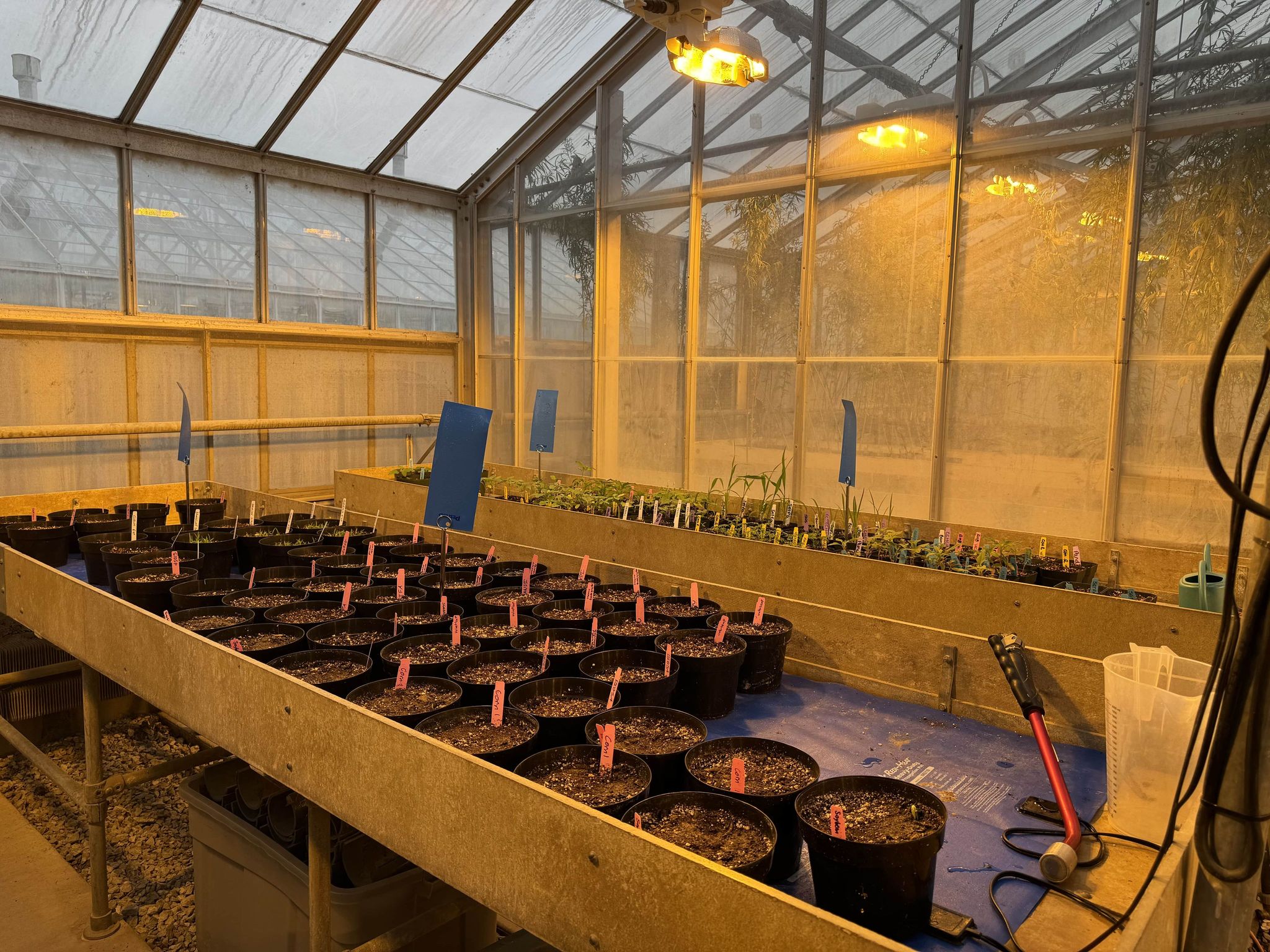}
    \vspace{-0.4cm}
    \caption{Greenhouse environment with lighting, temperature, and watering setup.}
    \label{fig:fig2_greenhouse}
    \vspace{-0.3cm}
\end{figure}

In this research, we conducted a study on 16 weed species at the SIU Horticulture Research Center greenhouse.
We began by preparing soil for seed planting, as shown in  Figure \ref{fig:fig1}(b). Potting soil (Pro-Mix ® BX) was placed into 32 square pots (10.7 cm x 10.7 cm x 9 cm), each labeled by species with white pot stakes. Two seeds from each species were planted per pot. Environmental conditions in the greenhouse, including temperature and lighting, were carefully controlled. Plants were watered as needed and fertilized with all-purpose 20-20-20 nutrient solution every 3 days. Figure \ref{fig:fig2_greenhouse} provides an overview of the greenhouse environment. We monitored the growth stages of each plant on a weekly basis, capturing images from the first week until week 11. Image capture ceased when the weeds entered their flowering stage, which marked the final growth phase in our study. We have captured our images by using an iPhone 15 Pro Max. Table \ref{species_frames} provides a comprehensive overview of our study, detailing the weed species codes, their corresponding scientific and common names, and the number of frames captured for each species on a weekly basis.

\begin{table*}[htbp]
\caption{Overview of Weed Species of Economic Concern, Corresponding Codes, and Weekly Frame Counts Captured for Each Species Across 11 Weeks in the Greenhouse}
\vspace{-0.4cm}
\label{species_frames}
\begin{center}
\resizebox{\textwidth}{!}{%
\begin{tabular}{|c|c|c|c|c|c|c|c|c|c|c|c|c|c|c|c|c|}
\hline
\multirow{2}{*}{\textbf{Species Code~\cite{kotleba1994european}}} & \multirow{2}{*}{\textbf{Scientific Name}~\cite{borsch2020world}} & \multirow{2}{*}{\textbf{Common Name}~\cite{wssaCompositeList}} & \multirow{2}{*}{\textbf{Family}} & \multirow{2}{*}{\textbf{Total Frames}} & \multicolumn{11}{|c|}{\textbf{Number of frames/week}} \\
\cline{6-16}
 &  &  &  &  & \textbf{\textit{W\_1}} & \textbf{\textit{W\_2}} & \textbf{\textit{W\_3}} & \textbf{\textit{W\_4}} & \textbf{\textit{W\_5}} & \textbf{\textit{W\_6}} & \textbf{\textit{W\_7}} & \textbf{\textit{W\_8}} & \textbf{\textit{W\_9}} & \textbf{\textit{W\_10}} & \textbf{\textit{W\_11}} \\
\hline
ABUTH & \textit{Abutilon theophrasti} Medik. & Velvetleaf & Malvaceae & 14754 & 1084 & 2451 & 1212 & 1819 & 1414 & 981 & 677 & 1164 & 1084 & 1500 & 1368 \\
AMAPA & \textit{Amaranthus palmeri} S. Watson. & Palmer Amaranth & Amaranthaceae & 17525 & 1441 & 1408 & 2110 & 2014 & 2441 & 1290 & 923 & 1478 & 1393 & 1667 & 1360 \\
AMARE & \textit{Amaranthus retroflexus} L. & Redroot Pigweed & Amaranthaceae & 15380 & 1017 & 1363 & 2110 & 1923 & 1884 & 1150 & 736 & 1237 & 1082 & 1596 & 1282 \\
AMATU & \textit{Amaranthus tuberculatus} (Moq.) Sauer. & Water Hemp & Amaranthaceae & 14852 & 1325 & 1459 & 1565 & 1664 & 1942 & 837 & 730 & 969 & 1638 & 1573 & 1150 \\
AMBEL & \textit{Ambrosia artemisiifolia} L. & Common Ragweed & Asteraceae & 17427 & 1022 & 2215 & 1846 & 1739 & 2162 & 1093 & 1066 & 1432 & 1092 & 2045 & 1715 \\
CHEAL & \textit{Chenopodium album} L. & Common Lambsquarter & Chenopodiaceae & 8015 & 1108 & 954 & 1416 & 661 & 1056 & 305 & 418 & 641 & 453 & 429 & 574 \\
CYPES & \textit{Cyperus esculentus} L. & Yellow Nutsedge & Cyperaceae & 14275 & 909 & 1512 & 1032 & 1499 & 2273 & 978 & 1224 & 1391 & 1182 & 1170 & 1105 \\
DIGSA & \textit{Digitaria sanguinalis} (L.) Scop. & Large Crabgrass & Poaceae & 16962 & 732 & 1312 & 2411 & 2596 & 1649 & 1335 & 1166 & 1261 & 1120 & 1628 & 1692 \\
ECHCG & \textit{Echinochloa crus-galli} (L.) P. Beauv. & Barnyard Grass & Poaceae & 16564 & 1349 & 2067 & 2029 & 1426 & 2221 & 1240 & 929 & 1280 & 1371 & 1332 & 1320 \\
ERICA & \textit{Erigeron canadensis} L. & Horse Weed & Asteraceae & 15134 & 930 & 2183 & 1691 & 1542 & 2715 & 1189 & 609 & 742 & 915 & 1217 & 1401 \\
PANDI & \textit{Panicum dichotomiflorum} Michx. & Full Panicum & Poaceae & 15182 & 1198 & 1400 & 2143 & 1296 & 1979 & 952 & 887 & 1350 & 1425 & 1034 & 1518 \\
SETFA & \textit{Setaria faberi} Herrm. & Gaint Foxtail & Poaceae & 14635 & 1614 & 1195 & 2083 & 1348 & 1944 & 1091 & 715 & 1466 & 843 & 1342 & 994 \\
SETPU & \textit{Setaria pumila} (Poir.) Roem. & Yellow Foxtail & Poaceae & 15211 & 887 & 1390 & 1732 & 1654 & 2040 & 1093 & 747 & 1361 & 1325 & 1348 & 1634 \\
SIDSP & \textit{Sida spinosa} L. & Princkly Sida & Malvaceae & 14452 & 1035 & 1782 & 1583 & 1259 & 2142 & 1373 & 804 & 1059 & 1186 & 1303 & 926 \\
SORHA & \textit{Sorghum halepense} (L.) Pers. & Johnson Grass & Poaceae & 10958 & 0 & 0 & 1444 & 1268 & 1395 & 945 & 749 & 1215 & 1328 & 1116 & 1498 \\
SORVU & \textit{Sorghum bicolor}  (L.) Moench. & Shatter Cane & Poaceae & 9573 & 945 & 1340 & 1959 & 832 & 1065 & 525 & 279 & 748 & 714 & 592 & 574 \\
\hline
\end{tabular}
}
\end{center}
\end{table*}



\begin{figure}[t]
    \centering
    \includegraphics[width=0.5\textwidth]{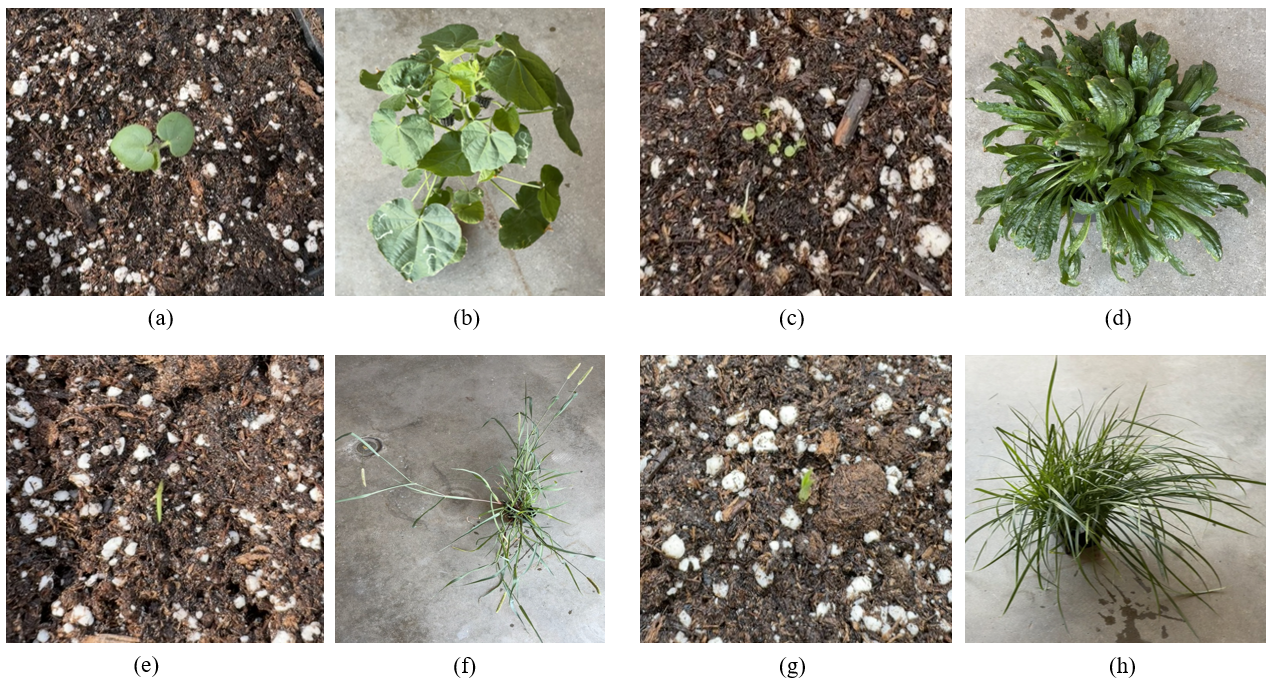}
    \vspace{-0.4cm}
    \caption{Growth stages example of four weed species. (a,b) ABUTH in week 1 and 11; (c,d) ERICA in week 1 and 11; (e,f) SETFA in week 1 and 11; (g,h) CYPES in week 1 and 11. Images show progression from seedling emergence to mature plants across different species.}
    \label{fig:fig3_plants}
    \vspace{-0.3cm}
\end{figure}

Among the 16 species of weeds studied, SORHA did not emerge in weeks 1 and 2. Consequently, the research encompassed a total of 174 classes. The full dataset initially comprised 2,494,476 frames. After a thorough review process to remove substandard images, 203,567 images were ultimately selected for training. Figure \ref{fig:fig3_plants} presents sample images of four weed species at different growth stages. For ABUTH, images from week 1 (a) and week 11 (b) are shown. Similarly, ERICA is represented by its week 1 (c) and week 11 (d) images. SETFA is depicted in its first week (e) and eleventh week (f) of growth. Lastly, CYPES is illustrated in its initial (g) and final (h) weeks of the study period. Notably, while several species produced flowers in their final growth stages, others did not, reflecting natural growth processes and photoperiod sensitivities.

\subsection{Data Preprocessing and Augmentation}
Our preprocessing pipeline begins with image normalization, a fundamental step that standardizes the input data. Each image is meticulously scaled to a 0-1 range by dividing all pixel values by 255.0. This normalization process is crucial as it ensures consistency across the dataset and aligns with the input requirements of neural networks, facilitating more efficient and effective training \cite{huang2023normalization}. Following normalization, we perform a color space conversion, transforming the images from the standard RGB (Red, Green, Blue) color space to the HSV (Hue, Saturation, Value) color space. The HSV color space allows us to more precisely isolate plant areas from the background, enhancing the accuracy of subsequent processing steps.

\begin{figure}[t]
    \centering
    \includegraphics[width=0.5\textwidth]{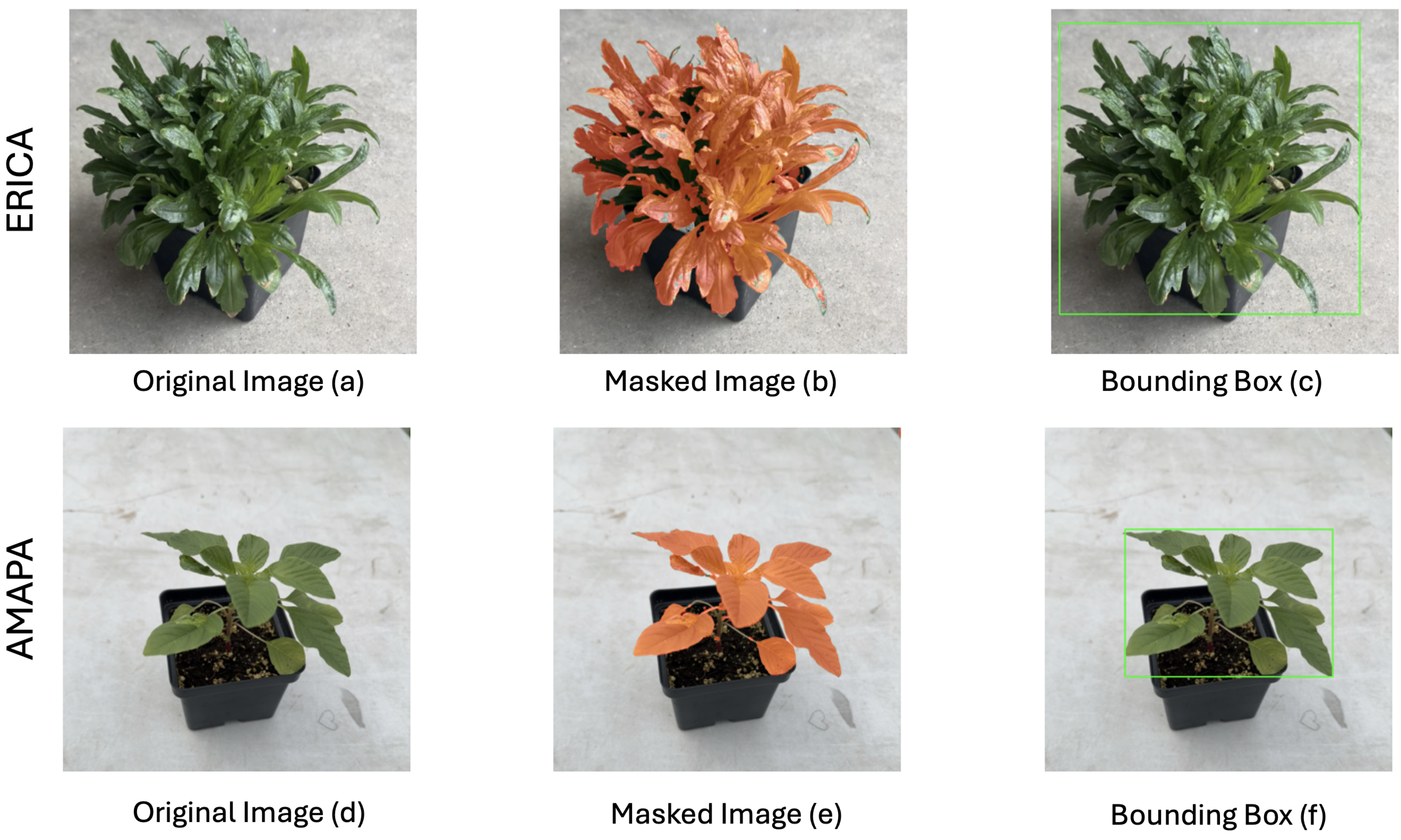}
    \vspace{-0.4cm}
    \caption{Data Augmentation process with original image, masked image, and bounding box, respectively, for ERICA (a,b,c) and AMAPA (d,e,f).}
    \label{fig:dataaug}
    \vspace{-0.3cm}
\end{figure}

The next step in our pipeline is green area detection. We employ carefully calibrated thresholds for the HSV channels to create a mask that highlights potential plant regions. Specifically, we use hue values ranging from 25/360 to 160/360, a minimum saturation value of 0.20. These thresholds have been empirically determined to effectively isolate green regions corresponding to plant matter while minimizing false positives from non-plant green objects. We apply morphological operations \cite{comer1999morphological} to refine the green mask and improve the continuity of detected plant areas. The refined green areas are then subjected to connected component analysis, which identifies and labels distinct regions within the image. This step is crucial for differentiating individual plants or plant clusters, allowing for more precise analysis and annotation. Fig \ref{fig:dataaug} shows the process of the data augmentation.

\subsection{Data Labeling}

Our labeling process creates comprehensive annotations for detected plants, including bounding box coordinates and detailed Pascal VOC XML annotations. We use Python libraries like Pillow \cite{clark2015pillow}, NumPy, and scikit-image for image processing. To ensure accuracy, we implemented a rigorous quality control process, manually refining annotations using LabelImg software \cite{labelImg2024} when necessary. Our labeling convention includes both species code and week number, enhancing the dataset's utility for tracking plant development and species-specific analysis. This meticulous approach results in a high-quality dataset with precise annotations and consistent formatting, suitable for various plant analysis tasks and growth stage tracking. 

\begin{figure}[t]
    \centering
    \includegraphics[width=0.5\textwidth]{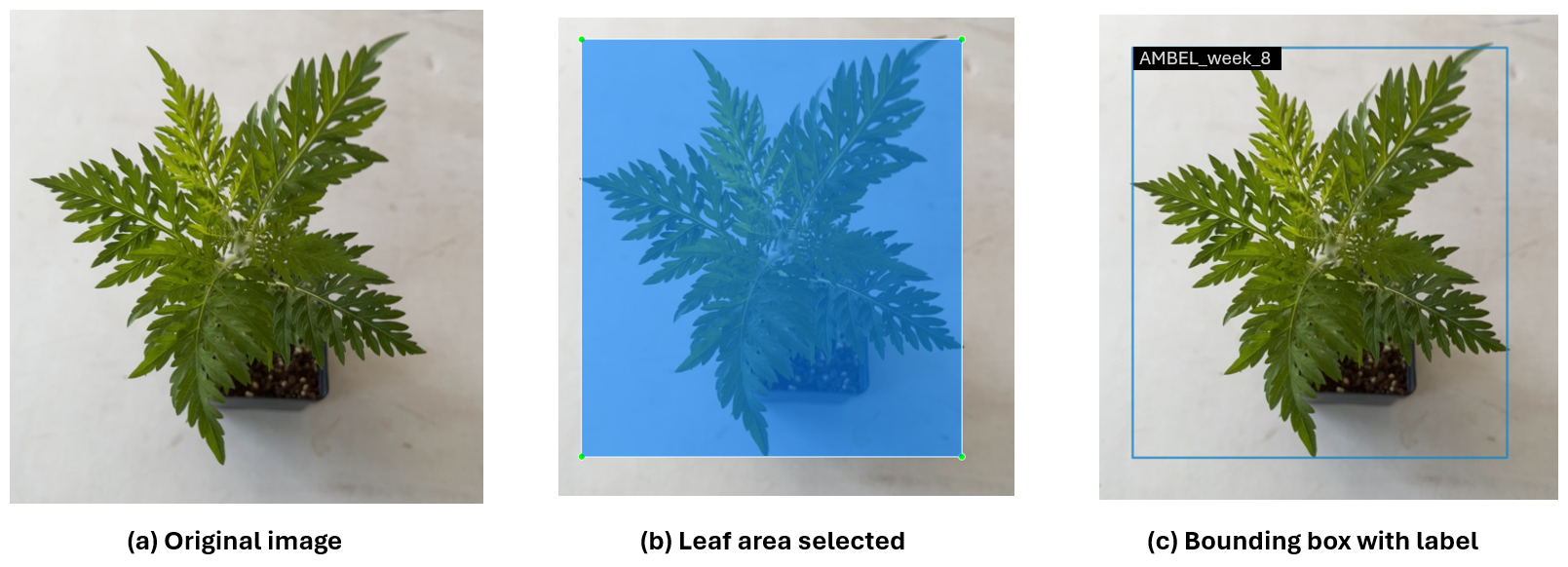}
    \vspace{-0.4cm}
    \caption{Illustration of the labeling process for weed detection. The original image (a) shows the weed plant, followed by the selected leaf area (b), highlighted in blue, and the final image (c) with a bounding box and label (AMBEL\_week\_8).}
    \label{fig:boundingbox}
    \vspace{-0.3cm}
\end{figure}

Figure \ref{fig:boundingbox} illustrates this process, presenting a side-by-side comparison of an original image and its corresponding labeled version, which we refer to as the ground truth.

\section{Methodology}

After annotating the dataset, we split the dataset into training, validation, and test sets. We used 184,719 images ($\sim$80\%) to train our object detection models and 23,090 images ($\sim$10\%) to validate the model during training time. The rest of the 23,090 images ($\sim$10\%) are held out to test the trained model’s performance. In this study, we employed two advanced deep-learning models for weed detection and classification: RetinaNet with a ResNeXt-101 backbone and Detection Transformer (DETR) with a ResNet-50 backbone. These models were tasked with classifying weed species and their respective growth stages (in weeks), while simultaneously localizing them within the images via bounding box predictions. We configured and trained these models using PyTorch and mmDetection on an NVIDIA RTX 3090 GPU.

\subsection{Detection Transformer with ResNet-50}

The Detection Transformer (DETR) model is an end-to-end object detection architecture that combines a convolutional backbone with a transformer encoder-decoder \cite{carion2020end}. This approach effectively addresses the complexities of identifying weeds in agricultural images. The backbone of our model ResNet-50 is a convolutional neural network, pre-trained on ImageNet (\texttt{open-mmlab://resnet50}). This 50-layer network, organized into four stages, serves as a powerful feature extractor. We utilize the output from the final stage (out\_indices=(3,)) and freeze the initial stages during training to preserve pre-learned features. The backbone's output can be represented as:
\vspace{-0.2cm}
\begin{equation}
F_{\text{resnet}} = \text{ResNet50}(I)
\end{equation}
where \(I\) is the input image. A Channel Mapper follows the backbone, transforming ResNet-50's 2048-channel output into a 256-channel feature map suitable for the transformer. This dimensionality reduction is achieved through a 1x1 convolution:
\vspace{-0.2cm}
\begin{equation}
F_{\text{neck}} = \text{Conv1x1}(F_{\text{resnet}})
\end{equation}

The core of our DETR model is the transformer module, comprising a 6-layer encoder and decoder. Each encoder layer incorporates a self-attention mechanism with 8 heads, followed by a feed-forward network (FFN) with ReLU activation. The model's bounding box head processes the decoder's output to predict class labels and bounding boxes. We employ cross-entropy loss for classification and a combination of L1 and Generalized IoU losses for bounding box regression. The overall loss function \cite{yin2019context} is defined as:
\vspace{-0.2cm}
\begin{equation}
L = \alpha \cdot L_{\text{cls}} + \beta \cdot L_{\text{bbox}} + \gamma \cdot L_{\text{iou}}
\end{equation}
where \(\alpha\), \(\beta\), and \(\gamma\) are weight coefficients. \( L_{\text{cls}} \) represents the classification loss, which in this case is the cross-entropy loss. \( L_{\text{bbox}} \) represents the bounding box regression loss, which is a combination of L1 loss and Generalized IoU loss, and \( L_{\text{iou}} \) 	represents the IoU loss, which is specifically aimed at improving the localization accuracy by penalizing the model based on the intersection over union between the predicted and ground truth bounding boxes. During training, we utilize the Hungarian algorithm \cite{ye2020cost} for bipartite matching, ensuring a one-to-one correspondence between predicted and ground-truth boxes. This approach optimizes the model's ability to accurately locate and classify weeds within agricultural images. By integrating the robust feature extraction capabilities of ResNet-50 with the DETR architecture's powerful attention mechanisms, our model achieves good performance in weed detection with 174 classes.

\subsection{RetinaNet with ResNeXt-101}

RetinaNet is a single-stage object detection model designed to address the extreme foreground-background class imbalance encountered during training \cite{li2019light}. The architecture comprises three main components: a backbone network for feature extraction, a neck (FPN) for generating multi-scale feature maps, and a detection head for predicting bounding boxes and class probabilities. We utilized ResNeXt-101 as the backbone, a variant of the ResNet architecture that employs grouped convolutions for improved efficiency and performance. The ResNeXt-101 backbone consists of 101 layers organized into four stages, with 32 groups and a base width of 4 channels per group. We initialized the backbone with weights pretrained on ImageNet (\texttt{open-mmlab://resnext101\_32x4d}) to leverage transfer learning. Batch normalization is applied after each convolutional layer to stabilize the learning process.

The Feature Pyramid Network (FPN) enhances the backbone's feature maps by combining high-level semantic features with low-level detailed features, enabling the detection of objects at various scales. The FPN generates multiple feature maps of different resolutions, which are then fed into the detection head. The detection head of RetinaNet comprises two subnetworks: a classification subnetwork for predicting object presence probabilities and a regression subnetwork for predicting bounding box coordinates. Each subnetwork consists of four convolutional layers, followed by a final convolutional layer that produces the desired outputs. To handle class imbalance, we employed the focal loss function \cite{lin2017focal} for training the classification subnetwork:
\vspace{-0.2cm}
\begin{equation}
\text{FL}(p_t) = -\alpha_t (1 - p_t)^\gamma \log(p_t)
\end{equation}

where \(p_t\) is the predicted probability, \(\alpha_t\) is a balancing factor, and \(\gamma\) is the focusing parameter.

We trained our model using an epoch-based training loop with the AdamW optimizer (learning rate \(lr = 0.0001\), weight decay \(wd = 0.0001\)). The learning rate schedule incorporated a linear warmup over the first 1000 iterations. We trained for 12 epochs with a batch size of 16, employing automatic learning rate scaling to accommodate potential batch size changes.

\subsection{Evaluation Metrics}

To assess the performance of our weed detection models, we employ a comprehensive set of metrics that capture both the accuracy and robustness of the detections. Our primary metrics are Average Precision (AP), Average Recall (AR), and Mean Average Precision (mAP) evaluated across various Intersection over Union (IoU) thresholds.

AP provides a single-value summary of the precision-recall curve, effectively balancing the trade-off between precision and recall. Precision (P) is defined as the ratio of true positive detections to the sum of true positive and false positive detections: $\text{} P = \frac{TP}{TP + FP}$ and Recall (R) is the ratio of true positive detections to the sum of true positive and false negative detections: $\text{} (R) = \frac{TP}{TP + FN}$.

In this research, a true positive is a detected bounding box that correctly identifies a weed species and has an IoU above a specified threshold (e.g., 0.50) with the ground truth bounding box. A false positive is a detection that either does not sufficiently overlap with any ground truth box or incorrectly identifies the weed species. A false negative occurs when a ground truth weed instance is not detected by the model. AP \cite{robertson2008new} is calculated by integrating the precision over the recall range and it can be defined as:
\vspace{-0.2cm}
\begin{equation}
\text{AP} = \int_0^1 P(R) \, dR
\end{equation}

AR \cite{zhu2004recall} measures the model's ability to detect all relevant objects. It is computed as the average of maximum recalls at specified IoU thresholds:
\vspace{-0.2cm}
\begin{equation}
\text{AR} = \frac{1}{N} \sum_{i=1}^N R_{\text{max}}(IoU_i)
\end{equation}

mAP is the mean of AP values across different classes and is a common metric for evaluating object detection models. It provides a balanced measure of precision and recall across various IoU thresholds. It can be defined as:
\vspace{-0.2cm}
\begin{equation}
\text{mAP} = \frac{1}{C} \sum_{c=1}^C AP_c
\end{equation}

where $AP_c$ is the Average Precision for class $c$, and $C$ is the total number of classes.
We evaluate these metrics at various IoU thresholds. This multi-faceted evaluation approach allows us to comprehensively analyze our models' capabilities in detecting and classifying weeds across various scenarios, providing insights into their precision, recall, and overall detection performance.

\section{Experimental Evaluation}

The evaluation encompasses both training and test datasets, with a detailed analysis across 16 weed species. We employ various metrics, including AP, AR at different IoU thresholds and detection limits, as well as mAP and mean average recall (mAR). Additionally, we compare the inference speed of both models to provide a holistic view of their capabilities.


\begin{table*}[t]
\vspace{-0.4cm}
\caption{Performance Comparison of DETR and RetinaNet on Training and Test Sets}
\label{performance_comparison}
\begin{center}
\begin{tabular}{|l|c|c|c|c|c|}
\hline
\multirow{2}{*}{\textbf{Model}} & \multicolumn{2}{c|}{\textbf{mAP}} & \multicolumn{2}{c|}{\textbf{mAR}} & \multirow{2}{*}{\textbf{FPS}} \\
\cline{2-5}
& \textbf{\textit{Train}} & \textbf{\textit{Test}} & \textbf{\textit{Train}} & \textbf{\textit{Test}} & \\
\hline
DETR & 0.854 & 0.840 & 0.941 & 0.936 & 3.49 \\
RetinaNet & \textbf{0.907} & 0.904 & \textbf{0.997} & 0.989 & \textbf{7.28} \\
\hline
\end{tabular}
\end{center}
\end{table*}

Table \ref{performance_comparison} compares DETR and RetinaNet performance on training and test sets, highlighting key metrics. RetinaNet consistently outperforms DETR across all presented metrics. In terms of mean Average Precision (mAP), RetinaNet achieves superior scores of 0.907 and 0.904 on training and test sets respectively, compared to DETR's 0.854 and 0.840. This trend continues in mean Average Recall (mAR), where RetinaNet approaches near-perfect scores with 0.997 (training) and 0.989 (test), while DETR achieves 0.941 and 0.936. Notably, RetinaNet's inference speed is significantly faster, operating at 7.28 Frames Per Second (FPS), more than twice the speed of DETR's 3.49 FPS. This substantial difference in processing speed, combined with RetinaNet's superior accuracy metrics, suggests it may be the more efficient choice for real-time or high-volume weed detection tasks.

\begin{table*}[t]
\caption{\footnotesize Performance Comparison of DETR and RetinaNet across Weed Species}
\label{weed_species_performance}
\vspace{-0.6cm}
\begin{center}
\resizebox{\textwidth}{!}{%
\begin{tabular}{|l|c|c|c|c|c|c|c|c|}
\hline
\multirow{2}{*}{\textbf{Species Code}} & \multicolumn{4}{c|}{\textbf{DETR}} & \multicolumn{4}{c|}{\textbf{RetinaNet}} \\
\cline{2-9}
& \textbf{\textit{Average mAP}} & \textbf{\textit{Average mAP\_50}} & \textbf{\textit{Average mAP\_75}} & \textbf{\textit{Average Recall}} & \textbf{\textit{Average mAP}} & \textbf{\textit{Average mAP\_50}} & \textbf{\textit{Average mAP\_75}} & \textbf{\textit{Average Recall}} \\
\hline
ABUTH & 0.683 & 0.907 & 0.719 & 0.973 & 0.720 & 0.924 & 0.779 & 0.993 \\
AMAPA & 0.617 & 0.835 & 0.672 & 0.975 & 0.877 & 0.985 & 0.939 & 0.994 \\
AMARE & 0.575 & 0.807 & 0.598 & 0.957 & 0.617 & 0.941 & 0.684 & 0.987 \\
AMATA & 0.536 & 0.721 & 0.565 & 0.869 & 0.832 & 0.977 & 0.905 & 0.997 \\
AMBEL & 0.817 & 0.978 & 0.898 & 0.993 & 0.663 & 0.926 & 0.740 & 0.994 \\
CHEAL & 0.503 & 0.846 & 0.502 & 0.962 & 0.871 & 0.993 & 0.957 & 0.997 \\
CYPES & 0.643 & 0.861 & 0.680 & 0.986 & 0.781 & 0.971 & 0.853 & 0.995 \\
DIGSA & 0.578 & 0.864 & 0.594 & 0.995 & 0.664 & 0.878 & 0.753 & 0.976 \\
ECHCG & 0.655 & 0.899 & 0.715 & 0.986 & 0.566 & 0.814 & 0.612 & 0.950 \\
ERICA & 0.718 & 0.918 & 0.752 & 0.977 & 0.678 & 0.918 & 0.749 & 0.992 \\
PANDI & 0.670 & 0.929 & 0.723 & 0.979 & 0.724 & 0.934 & 0.799 & 0.993 \\
SETFA & 0.680 & 0.903 & 0.756 & 0.990 & 0.785 & 0.967 & 0.854 & 0.993 \\
SETPU & 0.597 & 0.852 & 0.652 & 0.973 & 0.794 & 0.949 & 0.858 & 0.993 \\
SIDSP & 0.771 & 0.980 & 0.826 & 0.993 & 0.739 & 0.954 & 0.832 & 0.991 \\
SORVU & 0.582 & 0.791 & 0.624 & 0.871 & 0.713 & 0.925 & 0.789 & 0.995 \\
SORHA & 0.527 & 0.715 & 0.544 & 0.892 & 0.693 & 0.858 & 0.780 & 0.894 \\
\hline
\end{tabular}
}
\end{center}
\vspace{-0.5cm}
\end{table*}

Table \ref{weed_species_performance} delves deeper, breaking down performance across all individual weed species. This table shows the average value of all 11 weeks results for 16 species. This view reveals nuances in each model's capabilities. RetinaNet demonstrates more consistent performance across species, with less variation in mAP scores. In contrast, DETR's performance fluctuates more widely, excelling with some species like AMBEL (mAP 0.817) and SIDSP (mAP 0.771), while struggling with others such as CHEAL (mAP 0.503) and SORHA (mAP 0.527). RetinaNet shines particularly bright with species like AMATA (mAP 0.832) and AMAPA (mAP 0.877), though it faces challenges with ECHCG (mAP 0.566).
Across all species, RetinaNet consistently achieves higher recall, often nearing or reaching 1.0, while DETR's recall, though generally high, shows more variability. Both models exhibit the expected decline in mAP as the IoU threshold increases from 0.5 to 0.75, but RetinaNet maintains higher scores more consistently throughout this range.


We have selected four species for presenting their growth-wise experimental evaluation in this paper: Palmer amaranth (AMAPA), waterhemp (AMATA), giant foxtail (SETFA), and velvetleaf (ABUTH). These species are considered “driver weeds” or weeds that drive management decisions in USA agriculture due to their aggressive growth habits, herbicide resistance, and significant impact on crop yields \cite{illinois}. AMAPA and AMATA are particularly notorious for their rapid growth and resistance to multiple herbicide modes of action, making them difficult to control and highly competitive with crops.\\

\begin{table}
\caption{Performance Comparison of DETR and RetinaNet for SETFA}
\label{setfa_performance}
\vspace{-0.2cm}
\centering
\resizebox{\columnwidth}{!}{%
\begin{tabular}{|l|cccc|cccc|}
\hline
\multirow{2}{*}{\textbf{Class Name}} & \multicolumn{4}{c|}{\textbf{DETR}} & \multicolumn{4}{c|}{\textbf{RetinaNet}} \\
\cline{2-9}
& \textbf{\textit{mAP}} & \textbf{\textit{mAP\_50}} & \textbf{\textit{mAP\_75}} & \textbf{\textit{Recall}} & \textbf{\textit{mAP}} & \textbf{\textit{mAP\_50}} & \textbf{\textit{mAP\_75}} & \textbf{\textit{Recall}} \\
\hline
SETFA\_Week\_1 & 0.355 & 0.605 & 0.348 & 0.986 & 0.671 & 0.870 & 0.767 & 0.980 \\
SETFA\_Week\_2 & 0.400 & 0.763 & 0.414 & 1.000 & 0.623 & 0.801 & 0.738 & 0.989 \\
SETFA\_Week\_3 & 0.740 & 0.999 & 0.887 & 0.929 & 0.755 & 0.991 & 0.830 & 1.000 \\
SETFA\_Week\_4 & 0.607 & 0.860 & 0.717 & 0.974 & 0.555 & 0.764 & 0.611 & 1.000 \\
SETFA\_Week\_5 & 0.741 & 0.964 & 0.868 & 1.000 & 0.657 & 0.899 & 0.708 & 0.995 \\
SETFA\_Week\_6 & 0.658 & 0.859 & 0.669 & 1.000 & 0.648 & 0.936 & 0.682 & 1.000 \\
SETFA\_Week\_7 & 0.825 & 0.974 & 0.860 & 1.000 & 0.822 & 0.980 & 0.795 & 1.000 \\
SETFA\_Week\_8 & 0.743 & 1.000 & 0.818 & 1.000 & 0.822 & 1.000 & 0.943 & 1.000 \\
SETFA\_Week\_9 & 0.856 & 0.955 & 0.949 & 1.000 & 0.843 & 0.983 & 0.932 & 1.000 \\
SETFA\_Week\_10 & 0.696 & 0.956 & 0.802 & 0.986 & 0.643 & 0.956 & 0.738 & 0.986 \\
SETFA\_Week\_11 & 0.859 & 1.000 & 0.988 & 1.000 & 0.808 & 1.000 & 0.938 & 1.000 \\
\hline
\end{tabular}
}
\end{table}

\begin{table}
\caption{Performance Comparison of DETR and RetinaNet for AMAPA}
\label{amapa_performance}
\vspace{-0.2cm}
\centering
\resizebox{\columnwidth}{!}{%
\begin{tabular}{|l|cccc|cccc|}
\hline
\multirow{2}{*}{\textbf{Class Name}} & \multicolumn{4}{c|}{\textbf{DETR}} & \multicolumn{4}{c|}{\textbf{RetinaNet}} \\
\cline{2-9}
& \textbf{\textit{mAP}} & \textbf{\textit{mAP\_50}} & \textbf{\textit{mAP\_75}} & \textbf{\textit{Recall}} & \textbf{\textit{mAP}} & \textbf{\textit{mAP\_50}} & \textbf{\textit{mAP\_75}} & \textbf{\textit{Recall}} \\
\hline
AMAPA\_Week\_1 & 0.096 & 0.345 & 0.035 & 1.000 & 0.481 & 0.729 & 0.585 & 0.949 \\
AMAPA\_Week\_2 & 0.277 & 0.518 & 0.263 & 1.000 & 0.771 & 0.974 & 0.808 & 1.000 \\
AMAPA\_Week\_3 & 0.354 & 0.718 & 0.354 & 0.925 & 0.636 & 0.933 & 0.657 & 0.995 \\
AMAPA\_Week\_4 & 0.505 & 0.860 & 0.501 & 0.837 & 0.860 & 1.000 & 0.988 & 1.000 \\
AMAPA\_Week\_5 & 0.576 & 0.855 & 0.670 & 0.983 & 0.711 & 0.887 & 0.735 & 1.000 \\
AMAPA\_Week\_6 & 0.839 & 1.000 & 0.930 & 0.991 & 0.860 & 0.986 & 0.917 & 1.000 \\
AMAPA\_Week\_7 & 0.809 & 0.982 & 0.912 & 0.996 & 0.896 & 0.980 & 0.974 & 0.989 \\
AMAPA\_Week\_8 & 0.766 & 0.985 & 0.882 & 1.000 & 0.835 & 1.000 & 0.955 & 1.000 \\
AMAPA\_Week\_9 & 0.796 & 0.934 & 0.865 & 1.000 & 0.836 & 0.945 & 0.865 & 0.994 \\
AMAPA\_Week\_10 & 0.852 & 0.986 & 0.981 & 1.000 & 0.846 & 1.000 & 0.962 & 1.000 \\
AMAPA\_Week\_11 & 0.912 & 1.000 & 1.000 & 1.000 & 0.902 & 1.000 & 1.000 & 1.000 \\
\hline
\end{tabular}
}
\end{table}


\vspace{-0.2cm}

\begin{table}
\caption{Performance Comparison of DETR and RetinaNet for ABUTH}
\label{abuth_performance}
\vspace{-0.2cm}
\centering
\resizebox{\columnwidth}{!}{%
\begin{tabular}{|l|cccc|cccc|}
\hline
\multirow{2}{*}{\textbf{Class Name}} & \multicolumn{4}{c|}{\textbf{DETR}} & \multicolumn{4}{c|}{\textbf{RetinaNet}} \\
\cline{2-9}
& \textbf{\textit{mAP}} & \textbf{\textit{mAP\_50}} & \textbf{\textit{mAP\_75}} & \textbf{\textit{Recall}} & \textbf{\textit{mAP}} & \textbf{\textit{mAP\_50}} & \textbf{\textit{mAP\_75}} & \textbf{\textit{Recall}} \\
\hline
ABUTH\_Week\_1 & 0.418 & 0.723 & 0.471 & 0.994 & 0.605 & 0.899 & 0.689 & 1.000 \\
ABUTH\_Week\_2 & 0.576 & 0.988 & 0.530 & 1.000 & 0.829 & 0.990 & 0.952 & 1.000 \\
ABUTH\_Week\_3 & 0.356 & 0.697 & 0.346 & 1.000 & 0.790 & 0.996 & 0.899 & 1.000 \\
ABUTH\_Week\_4 & 0.408 & 0.771 & 0.396 & 0.996 & 0.725 & 0.973 & 0.844 & 0.995 \\
ABUTH\_Week\_5 & 0.445 & 0.923 & 0.377 & 0.871 & 0.730 & 0.974 & 0.789 & 1.000 \\
ABUTH\_Week\_6 & 0.850 & 1.000 & 1.000 & 0.886 & 0.924 & 0.970 & 0.970 & 0.972 \\
ABUTH\_Week\_7 & 0.885 & 0.932 & 0.931 & 0.993 & 0.966 & 1.000 & 1.000 & 1.000 \\
ABUTH\_Week\_8 & 0.856 & 1.000 & 0.982 & 1.000 & 0.911 & 1.000 & 1.000 & 1.000 \\
ABUTH\_Week\_9 & 0.912 & 0.977 & 0.949 & 0.975 & 0.876 & 0.978 & 0.920 & 1.000 \\
ABUTH\_Week\_10 & 0.880 & 0.967 & 0.923 & 1.000 & 0.868 & 0.971 & 0.893 & 1.000 \\
ABUTH\_Week\_11 & 0.924 & 1.000 & 1.000 & 0.989 & 0.924 & 1.000 & 1.000 & 1.000 \\
\hline
\end{tabular}
}
\end{table}

\begin{table}
\caption{Performance Comparison of DETR and RetinaNet for AMATA}
\label{amata_performance}
\vspace{-0.2cm}
\centering
\resizebox{\columnwidth}{!}{%
\begin{tabular}{|l|cccc|cccc|}
\hline
\multirow{2}{*}{\textbf{Class Name}} & \multicolumn{4}{c|}{\textbf{DETR}} & \multicolumn{4}{c|}{\textbf{RetinaNet}} \\
\cline{2-9}
& \textbf{\textit{mAP}} & \textbf{\textit{mAP\_50}} & \textbf{\textit{mAP\_75}} & \textbf{\textit{Recall}} & \textbf{\textit{mAP}} & \textbf{\textit{mAP\_50}} & \textbf{\textit{mAP\_75}} & \textbf{\textit{Recall}} \\
\hline
AMATA\_Week\_1 & 0.001 & 0.003 & 0.000 & 0.982 & 0.641 & 0.981 & 0.742 & 0.992 \\
AMATA\_Week\_2 & 0.004 & 0.021 & 0.000 & 1.000 & 0.529 & 0.923 & 0.525 & 0.966 \\
AMATA\_Week\_3 & 0.157 & 0.397 & 0.076 & 0.391 & 0.747 & 0.998 & 0.934 & 1.000 \\
AMATA\_Week\_4 & 0.484 & 0.910 & 0.486 & 0.397 & 0.763 & 0.985 & 0.838 & 1.000 \\
AMATA\_Week\_5 & 0.544 & 0.974 & 0.541 & 0.839 & 0.738 & 0.961 & 0.822 & 0.994 \\
AMATA\_Week\_6 & 0.763 & 0.960 & 0.878 & 0.970 & 0.923 & 0.994 & 0.972 & 1.000 \\
AMATA\_Week\_7 & 0.905 & 1.000 & 0.977 & 0.995 & 0.968 & 1.000 & 0.974 & 1.000 \\
AMATA\_Week\_8 & 0.756 & 0.913 & 0.808 & 1.000 & 0.889 & 0.979 & 0.954 & 0.990 \\
AMATA\_Week\_9 & 0.881 & 0.960 & 0.952 & 1.000 & 0.926 & 0.990 & 0.972 & 1.000 \\
AMATA\_Week\_10 & 0.520 & 0.797 & 0.529 & 0.989 & 0.625 & 0.927 & 0.670 & 1.000 \\
AMATA\_Week\_11 & 0.882 & 0.993 & 0.965 & 1.000 & 0.849 & 0.998 & 0.933 & 1.000 \\
\hline
\end{tabular}
}
\end{table}

\begin{figure}[t]
    \centering
    \includegraphics[width=0.5\textwidth]{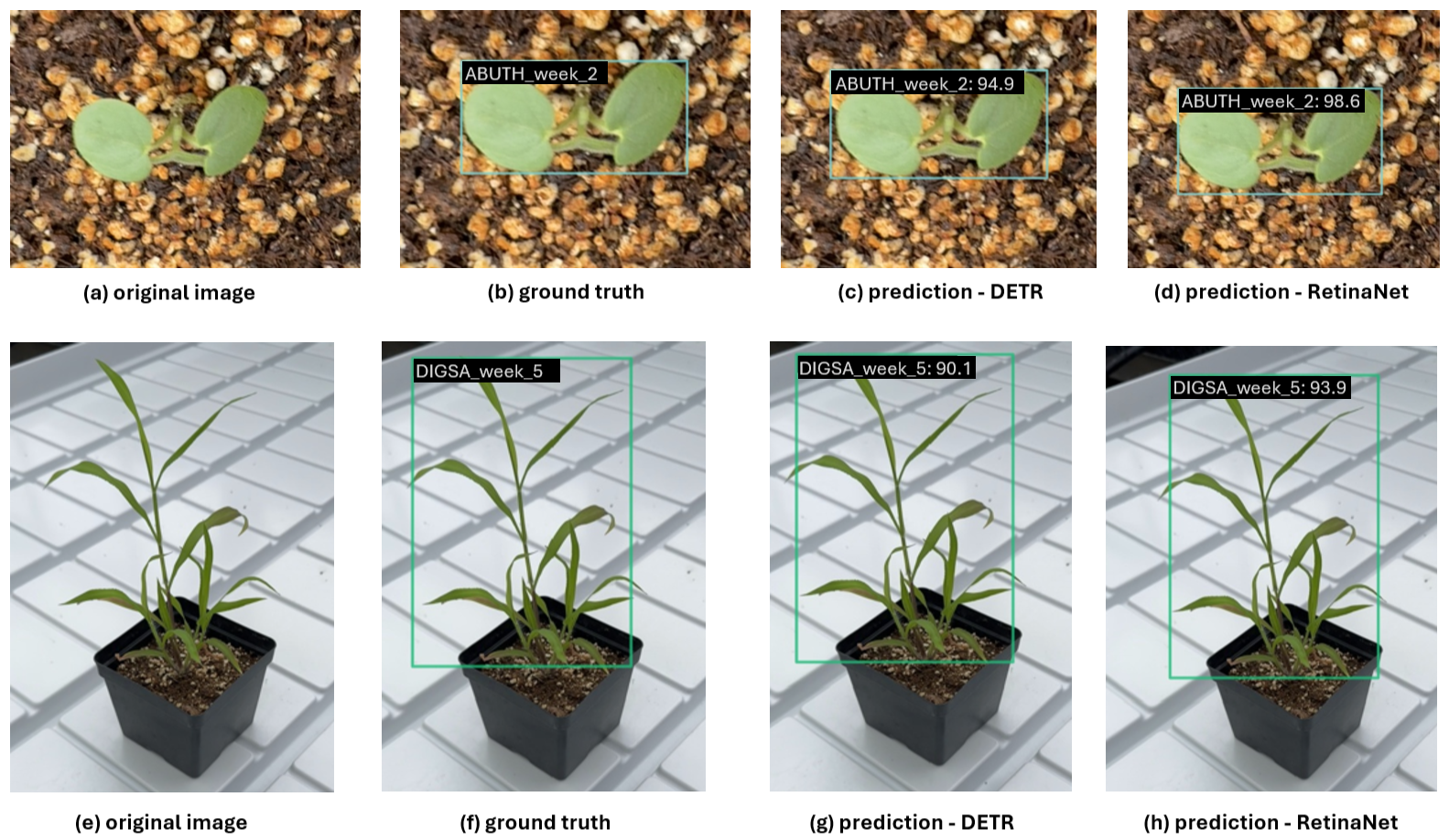}
    \vspace{-0.4cm}
    \caption{Comparison of object detection results for ABUTH and DIGSA using DETR and RetinaNet models. Row 1 displays predictions for ABUTH, and Row 2 displays predictions for DIGSA, with ground truth and model confidence scores indicated for each detection.}
    \label{fig:detectionresult}
    \vspace{-0.3cm}
\end{figure}

Tables \ref{setfa_performance}, \ref{amapa_performance}, \ref{abuth_performance}, and \ref{amata_performance} present comprehensive performance comparisons between DETR and RetinaNet across four weed species (SETFA, AMAPA, ABUTH, and AMATA) over 11 weeks. Both models demonstrated high performance across various metrics, including mAP, mAP\_50, mAP\_75, and Recall. RetinaNet generally outperformed DETR, showing more consistent and often higher scores across most species and weeks. For instance, RetinaNet achieved peak mAP scores of 0.843 for SETFA, 0.902 for AMAPA, 0.924 for ABUTH, and 0.968 for AMATA. DETR's highest mAP scores were comparable, reaching 0.859 for SETFA, 0.912 for AMAPA, 0.924 for ABUTH, and 0.905 for AMATA. Both models frequently achieved perfect scores of 1.000 in mAP\_50 and Recall metrics across various weeks and species, indicating excellent detection accuracy at lower IoU thresholds and high object detection rates.\\
However, both models exhibited some performance fluctuations, particularly in the early weeks. DETR often struggled more in the initial weeks, with notably low mAP scores such as 0.355 for SETFA in Week 1, 0.096 for AMAPA in Week 1, and 0.001 for AMATA in Week 1. RetinaNet generally showed more stability, with its lowest mAP scores being higher than DETR's in most cases. For example, RetinaNet's lowest mAP for SETFA was 0.555 in Week 4, for AMAPA it was 0.481 in Week 1, and for AMATA it was 0.529 in Week 2. These early-week challenges could be attributed to factors such as less robust feature extraction, difficulty in detecting small objects, or lower-quality images in the initial stages of plant growth. Despite these early challenges, both models demonstrated significant improvement over time, with peak performances often occurring in later weeks (Weeks 8-11). This trend suggests that as plants matured and image quality potentially improved, both DETR and RetinaNet were able to more accurately detect and classify the weed species.

Figure \ref{fig:detectionresult} shows the prediction result of DETR and RetinaNet model. The top row focuses on ABUTH, where the first image (a) shows the original plant without any annotations, followed by the (b) ground truth with a labeled bounding box indicating "ABUTH week 2." The subsequent images display predictions by (c) DETR and (d) RetinaNet models, each bounding box labeled with the species name, the corresponding week, and the model's confidence score, with RetinaNet showing a slightly higher score (98.6) compared to DETR (94.9). The bottom row repeats this structure for DIGSA, showing the (e) original image, (f) the ground truth ("DIGSA week 5"), and the predictions from (g) DETR and (h) RetinaNet. For DIGSA, the confidence scores are close, with DETR predicting 90.1 and RetinaNet predicting 93.9, both models accurately detecting the plant but with varying degrees of confidence.
\section{Conclusion}
This research marks a pivotal advancement in precision agriculture by demonstrating the effectiveness of AI models, particularly RetinaNet, in weed detection and classification across various growth stages and species. Our study, conducted on a comprehensive dataset of 203,567 images spanning 16 weed species over 11 weeks, reveals RetinaNet's superior performance with mAP scores of 0.907 and 0.904 on training and test sets, and an inference speed of 7.28 FPS, significantly outpacing DETR's 0.854 and 0.840 mAP scores and 3.49 FPS speed. Both models exhibit improved accuracy with plant maturation, yet challenges persist during the early growth stages (weeks 1-2) due to poor differentiation between emerging plants and soil. These findings underscore the practical implications for weed management, with RetinaNet recommended for real-time applications due to its accuracy and speed. To integrate these models into existing agricultural practices, farmers should implement mobile-based applications for in-field weed detection using RetinaNet, calibrate the model for specific weed species with their growth stages prevalent in their region, and combine AI-driven detection with GPS-guided precision spraying systems. Despite the controlled greenhouse setting and early-stage detection challenges, this study lays the groundwork for future research aimed at enhancing detection accuracy through custom transformer models and expanding the dataset to include real field conditions. These AI-driven innovations hold the promise of revolutionizing weed management by enabling species-specific, growth-stage-aware detection, potentially reducing herbicide use, cutting costs, and minimizing environmental impact. By following these integration guidelines, farmers can leverage AI models to optimize their weed management strategies, leading to more sustainable and efficient agricultural practices.





\end{document}